\documentclass[5p]{elsarticle}

\usepackage{lineno,hyperref}
\usepackage{amsmath,amssymb,amsfonts}
\usepackage[ruled,vlined]{algorithm2e}
\usepackage{booktabs}
\usepackage{caption}
\usepackage{capt-of}
\usepackage{comment}
\usepackage{color}
\usepackage{float}
\usepackage{graphicx}
\usepackage{multirow}
\usepackage{nameref}
\usepackage{setspace}
\usepackage{subfigure}
\usepackage{tabularx}
\usepackage{textcomp}
\usepackage{wrapfig}
\def\BibTeX{{\rm B\kern-.05em{\sc i\kern-.025em b}\kern-.08em
		T\kern-.1667em\lower.7ex\hbox{E}\kern-.125emX}}

\DeclareMathOperator*{\argmin}{arg\,min}

\journal{Journal of Pattern Recognition}

\begin{document}
	
	\begin{frontmatter}

	\title{Image Classification with Deep Learning in the Presence of Noisy Labels: A Survey}

	\author[1]{G\"orkem Algan\corref{cor1}\fnref{aselsan,metu}}
	\ead{e162565@metu.edu.tr}
	
	\author[2]{Ilkay Ulusoy\fnref{metu}}
	\ead{ilkay@metu.edu.tr}
	
	\cortext[cor1]{Corresponding author}
	\fntext[aselsan]{ASELSAN, Ankara}
	\fntext[metu]{Middle East Technical University, Electrical-Electronics Engineering, Ankara}
	\address[1]{ASELSAN, Balikhisar mah. Cankiri bulvari 7.km No:89, 06750, Ankara, Turkey}
	\address[2]{METU, Universiteler mah. Dumlupinar mah.Electrical-Electronics Engineering Department A-410, 06800, Ankara, Turkey}
	
	\begin{abstract}
        Image classification systems recently made a giant leap with the advancement of deep neural networks. However, these systems require an excessive amount of labeled data to be adequately trained. Gathering a correctly annotated dataset is not always feasible due to several factors, such as the expensiveness of the labeling process or difficulty of correctly classifying data, even for the experts. Because of these practical challenges, label noise is a common problem in real-world datasets, and numerous methods to train deep neural networks with label noise are proposed in the literature. Although deep neural networks are known to be relatively robust to label noise, their tendency to overfit data makes them vulnerable to memorizing even random noise. Therefore, it is crucial to consider the existence of label noise and develop counter algorithms to fade away its adverse effects to train deep neural networks efficiently. Even though an extensive survey of machine learning techniques under label noise exists, the literature lacks a comprehensive survey of methodologies centered explicitly around deep learning in the presence of noisy labels. This paper aims to present these algorithms while categorizing them into one of the two subgroups: noise model based and noise model free methods. Algorithms in the first group aim to estimate the noise structure and use this information to avoid the adverse effects of noisy labels. Differently, methods in the second group try to come up with inherently noise robust algorithms by using approaches like robust losses, regularizers or other learning paradigms.
	\end{abstract}
	

	\end{frontmatter}

	
	\section{Introduction} \label{sec:1}
Recent advancement in deep learning has led to great improvements on many different domains, such as image classification \cite{krizhevsky2012imagenet,he2016deep,simonyan2014very}, object detection \cite{girshick2014rich,ren2015faster,liu2016ssd}, semantic segmentation \cite{lin2016efficient,long2015fully} and others. Despite their impressive ability for representation learning \cite{rolnick2017deep,drory2018neural}, it is shown that these powerful models can overfit to even complete random noise \cite{zhang2016understanding}. Various works are devoted to explain this phenomenon \cite{krueger2017deep,arpit2017closer}, yet regularizing deep neural networks (DNNs) while avoiding overfitting stays to be an important challenge. It gets even more crucial when there exists noise in data. Therefore, various methods are proposed in the literature to train deep neural networks effectively in the presence of noise.

There are two kinds of noise in the literature: feature noise and label noise \cite{zhu2004class}. Feature noise corresponds to the corruption in observed data features, while label noise means the change of label from its actual class. Even though both noise types may cause a significant decrease in the performance \cite{frenay2014classification,frenay2014comprehensive}, label noise is considered to be more harmful \cite{zhu2004class, hataya2018investigating} and shown to deteriorate the performance of classification systems in a broad range of problems \cite{zhu2004class, nettleton2010study, pechenizkiy2006class}. This is due to several factors; the label is unique for each data while features are multiple, and the importance of each feature varies while the label always has a significant impact \cite{frenay2014classification}. This work focuses on label noise; therefore, noise and label noise is used synonymously throughout the article.

The necessity of an excessive amount of labeled data for supervised learning is a significant drawback since it requires an expensive dataset collection and labeling process. To overcome this issue, cheaper alternatives have emerged. For example, an almost unlimited amount of data can be collected from the web via search engines or social media. Similarly, the labeling process can be crowdsourced with the help of systems like Amazon Mechanical Turk\footnote{http://www.mturk.com}, Crowdflower\footnote{http://crowdflower.com}, which decrease the cost of labeling notably. Another widely used approach is to label data with automated systems. However, all these approaches led to a common problem; label noise. Besides these methods, label noise can occur even in the case of expert annotators. Labelers may lack the necessary experience, or data can be too complex to be correctly classified, even for the experts. Moreover, label noise can also be introduced to data for adversarial poisoning purposes \cite{li2016data,steinhardt2017certified}. Being a natural outcome of dataset collection and labeling process makes label noise robust algorithms an essential topic for the development of efficient computer vision systems.  

Supervised learning with label noise is an old phenomenon with three decades of history \cite{angluin1988learning}. An extensive survey about relatively old machine learning techniques under label noise is available \cite{frenay2014classification,frenay2014comprehensive}. However, no work is proposed to provide a comprehensive survey on classification methods centered around deep learning in the presence of label noise. This work focuses explicitly on filling this absence. Even though deep networks are considered to be relatively robust to label noise \cite{rolnick2017deep,drory2018neural}, they have an immense capacity to overfit data \cite{zhang2016understanding}. Therefore, preventing DNNs to overfit noisy data is very important, especially for fail-safe applications, such as automated medical diagnosis systems. Considering the significant success of deep learning over its alternatives, it is a topic of interest, and many works are presented in the literature. Throughout the paper, these methods are briefly explained and grouped to provide the reader with a clear overview of the literature.

This paper is organized as follows. Section \ref{preliminaries} explains several concepts that are used throughout the paper. Proposed solutions in literature are categorized into two major groups, and these methods are discussed in \autoref{noisemodelbased} - \autoref{noisemodelfree}. In \autoref{experiments} widely used experimental setups are presented, and leaderboard on a benchmarking dataset is provided. Finally, \autoref{conclusion} concludes the paper.

	\section{Preliminaries} \label{preliminaries}
This section introduces necessary concepts for a better understanding of the paper. Firstly, the problem statement for supervised learning in the presence of noisy labels is given. Secondly, types of label noises are presented. Finally, sources of label noise are discussed.

\subsection{Problem Statement}
Classical supervised learning consists of an input dataset $\mathcal{S}=\{(x_1,y_1),...,(x_N,y_N)\}\in (X,Y)^N$ drawn according to an unknown distribution $\mathcal{D}$, over $(X,Y)$. The learning objective is to find the best mapping function $f:X \rightarrow Y$ among family of functions $\mathcal{F}$, where each function is parametrized by $\theta$.

One way of evaluating the performance of a classifier is the so called loss function, denoted as $l:\mathcal{R}\times Y \rightarrow \mathcal{R^+}$. Given an example $(x_i,y_i) \in (X,Y)$, $l(f_\theta(x_i),y_i)$ evaluates how good is the classifier prediction. Then, for any classifier $f$, expected risk is defined as follow, where E denotes the expectation over distribution $\mathcal{D}$.

\begin{equation}
    R_{l,\mathcal{D}}(f_\theta)=E_\mathcal{D}[l(f_\theta(x),y)]
\end{equation}

Since it is not generally feasible to have complete knowledge over distribution $\mathcal{D}$, as an approximation, the empirical risk is used.

\begin{equation}
    \hat{R}_{l,\mathcal{D}}(f_\theta)=\dfrac{1}{N}\sum_{i=1}^{N}l(f_\theta(x_i),y_i)
\end{equation}

Various methods of learning a classifier may be seen as minimizing the empirical risk subjected to network parameters. 

\begin{equation}
    \theta^\star = \underset{\theta}{\argmin}\hat{R}_{l,\mathcal{D}}(f_\theta)
\end{equation}

In the presence of the label noise, dataset turns into $\mathcal{S}_n=\{(x_1,\tilde{y}_1),...,(x_N,\tilde{y}_N)\}\in (X,Y)^N$ drawn according to a noisy distribution $\mathcal{D}_n$, over $(X,Y)$. Then, the risk minimization results in as follows.

\begin{equation}
    \theta_n^\star = \underset{\theta}{\argmin}\hat{R}_{l,\mathcal{D}_n}(f_\theta)
\end{equation}

As a result, obtained parameters by minimizing over $\mathcal{D}_n$ are different from desired optimal classifier parameters.

\begin{center}
    $\theta^\star \neq \theta_n^\star$
\end{center}

Classical supervised learning aims to find the best estimator parameters $\theta^\star$ for given distribution $\mathcal{D}$ while iterating over $\mathcal{D}$. However, in noisy label setup, the task is still finding $\theta^\star$ while working on distribution $\mathcal{D}_n$. Therefore, classical risk minimization is insufficient in the presence of label noise since it would result in $\theta_n^\star$. As a result, variations of classical risk minimization methods are proposed in the literature, and they will be further evaluated in the upcoming sections.

\subsection{Label Noise Models} \label{labelnoisemodels}
A detailed taxonomy of label noise is provided in \cite{frenay2014classification}. In this work, we follow the same taxonomy with a little abuse of notation. Label noise can be affected by three factors: data features, the true label of data, and the labeler characteristics. According to the dependence of these factors, label noise can be categorized into three subclasses.

\begin{figure*}[h]
    \centering
    \includegraphics[width=\textwidth]{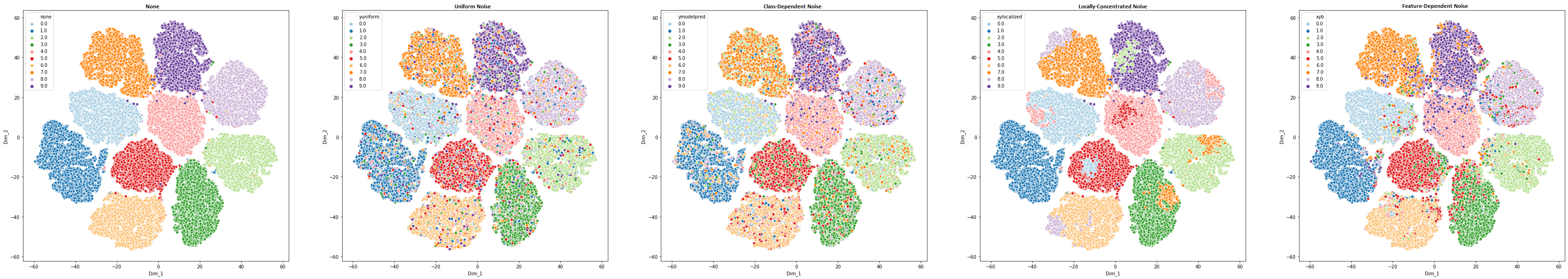}
    \caption{T-SNE plot of data distribution of MNIST dataset in feature space for 25\% noise ratio. a) clean data b) random noise c) y-dependent noise which is still randomly distributed in feature domain d) xy-dependent noise in locally concentrated form e) xy-dependent noise that is concentrated on decision boundaries }
    \label{fig:tsneplots}
\end{figure*}

\textit{Random noise} is totally random and depends on neither instance features nor its true class. With a given probability $p_e$ label is changed from its true class. \textit{Y-dependent noise} is independent of image features but depends on its class; $p_e = p(e|y)$. That means data from a particular class are more likely to be mislabeled. For example, in a handwritten digit recognition task, "3" and "8" are much more likely to be confused with each other rather than "3" and "5". \textit{XY-dependent noise} depends on both image features and its class; $p_e = p(e|x,y)$. As in the y-dependent case, objects from a particular class may be more likely to be mislabeled. Moreover, the chance of mislabeling may change according to data features. If an instance has similar features to another instance from another class, it is more likely to be mislabeled. Generating xy-dependent synthetic noise is harder than the previous two models; therefore, some works tried to provide a generic framework by either checking the complexity of data \cite{garcia2019new} or their position in feature space \cite{algan2020label}. All these types of noises are illustrated in \autoref{fig:tsneplots}

The case of multi-labeled data, in which each instance has multiple labels given by different annotators, is not considered here. In that scenario, works show that modeling each labeler's characteristics and using this information during training significantly boosts the performance \cite{guan2018said}. However, various characteristics of different labelers can be explained with given noise models. For example, in a crowd-sourced dataset, some labelers can be total spammers who label with a random selection \cite{khetan2017learning}; therefore, they can be modeled as random noise. On the other hand, labelers with better accuracies than random selection can be modeled by y-dependent or xy-dependent noise. As a result, the labeler's characteristic is not introduced as an extra ingredient in these definitions.

\subsection{Sources of Label Noise} \label{sourcesoflabelnoise}
As mentioned, label noise is a natural outcome of dataset collection process and can occur in various domains, such as medical imaging \cite{dgani2018training,xue2019robust,guan2018said}, semantic segmentation \cite{lu2016learning,zhu2018improving,acuna2019devil}, crowd-sourcing \cite{welinder2010multidimensional}, social network tagging \cite{cha2012social,wang2016sentiment}, financial analysis \cite{ait2010high} and many more. This work focuses on various solutions to such problems, but it may be helpful to investigate the causes of label noise to understand the phenomenon better.

Firstly, with the availability of the immense amount of data on the web and social media, it is a great interest of computer vision community to make use of that \cite{schroff2010harvesting,fergus2010learning,chen2013neil,divvala2014learning,joulin2016learning,krause2016unreasonable}. Nevertheless, labels of these data are coming from messy user tags or automated systems used by search engines. These processes of obtaining datasets are well known to result in noisy labels.

Secondly, the dataset can be labeled by multiple experts resulting in a multi-labeled dataset. Each labeler has a varying level of expertise, and their opinions may commonly conflict with each other, which results in noisy label problem \cite{khetan2017learning}. There are several reasons to get data labeled by more than one expert. Opinions of multiple labelers can be used to double-check each other's predictions for challenging datasets, or crowd-sourcing platforms can be used to decrease the cost of labeling for big data. Despite its cheapness, labels obtained from non-experts are commonly noisy with a differentiating rate of error. Some labelers even can be a total spammer who labels with random selection \cite{khetan2017learning}.

Thirdly, data can be too complicated for even the experts in the field, e.g., medical imaging. For example, to collect gold standard validation data for retinal images, annotations are gathered from 6-8 different experts \cite{de2018clinically,gulshan2016development}. This complexity can be due to the subjectiveness of the task for human experts or the lack of annotator experience. Considering the fields where the accurate diagnosis is of crucial importance, overcoming this noise is of great interest.

Lastly, label noise can intentionally be injected in purpose of regularizing \cite{xie2016disturblabel} or data poisoning \cite{li2016data,steinhardt2017certified}.

\subsection{Methodologies} \label{methodologies}
There are many possible ways to group proposed methods in the literature. For example, one possible way to distinguish algorithms is according to their need for a noise-free subset of data or not. Alternatively, they can be divided according to the noise type they are dealing with or label type such as singly-labeled or multi-labeled. However, these are not handy to understand the main approaches behind the proposed algorithms; therefore, different sectioning is proposed as noise model based and noise model free methods.

Noise model based methods aim to model the noise structure so that this information can be used during training to come through noisy labels. In general, approaches in this category aim to extract noise-free information contained within the dataset by either neglecting or de-emphasizing information coming from noisy samples. Furthermore, some methods attempt to reform the dataset by correcting noisy labels to increase the quality of the dataset for the classifier. The performance of these methods is heavily dependent on the accurate estimate of the underlying noise. The advantage of noise model based methods is the decoupling of classification and label noise estimation, which helps them to work with the classification algorithm at hand. Another good side is in the case of prior knowledge about the noise structure, noise model based methods can easily be head-started with this extra information inserted to the system.

Differently, noise model free methods aim to develop inherently noise robust strategies without explicit modeling of the noise structure. These approaches assume that the classifier is not too sensitive to the noise, and performance degradation results from overfitting. Therefore, the main focus is given to overfit avoidance by regularizing the network training procedure.

Both of the mentioned approaches are discussed and further categorized in \autoref{noisemodelbased} and \autoref{noisemodelfree}. \autoref{table:methods} presents all the mentioned methods to provide a clear picture as a whole. It should be noted that most of the time there are no sharp boundaries among the algorithms, and they may belong to more than one category. However, for the sake of integrity, they are placed in the subclass of most resemblance.

\begin{singlespace}
\begin{table*}[]
    \begin{tabular}{|l|l|}
    \hline
    \multirow{22}{*}{\rotatebox[origin=c]{90}{\textbf{\nameref{noisemodelbased}}}} & 
    \begin{tabular}[c]{@{}l@{}}\textbf{1. \nameref{noisychannel}}\\ 
        \textit{a.\nameref{noisychannelexplicit}}: predictions on noisy data \cite{patrini2017making}, predictions on clean data \cite{hendrycks2018using}\\
        easy data \cite{chen2015webly}\\
        \textit{b.\nameref{noisychanneliterative}}: EM \cite{bekker2016training,goldberger2016training,dgani2018training}, fully connected layer \cite{sukhbaatar2014training}, anchor point estimate \cite{xia2019anchor}\\
        Drichlet-distribution \cite{yao2019safeguarded}\\
        \textit{c.\nameref{noisychannelcomplex}}: noise type estimation \cite{xiao2015learning}, relevance estimation \cite{misra2016seeing}
    \end{tabular}\\\cline{2-2} 
    & \begin{tabular}[c]{@{}l@{}}\textbf{2. \nameref{labelnoisecleansing}}\\ 
        \textit{a.\nameref{labelcleansingclean}}: train on clean set \cite{lee2019photometric}, ensemble \cite{yuan2018iterative}, graph-based \cite{vahdat2017toward}\\
        \textit{b.\nameref{labelcleansingcleannoisy}}: iteratively correct \cite{veit2017learning}, correct for fine-tune \cite{dehghani2017fidelity}\\
        \textit{c.\nameref{labelcleansingnoisy}}: calculate posterior \cite{tanaka2018joint}, posterior with compatibility \cite{yi2019probabilistic}\\ consistency with model \cite{liu2017self,zheng2020error,arazo2019unsupervised}, ensemble \cite{zhang2017improving}, prototypes \cite{han2019deep}, quality embedding \cite{yao2018deep}\\
        partial labels \cite{durand2019learning}
    \end{tabular}\\\cline{2-2} 
    & \begin{tabular}[c]{@{}l@{}}\textbf{3. \nameref{datasetpruning}}\\ 
        \textit{a.Data pruning} network prediction based \cite{delany2012profiling}, ensemble of filters \cite{garcia2015using,luengo2018cnc}, according to noise rate \cite{northcutt2017learning}\\
        transfer learning \cite{wu2018light}, cyclic state \cite{huang2019o2u}, K-means \cite{sharma2020noiserank}\\
        \textit{b.Label pruning} semi-supervised learning \cite{ding2018semi,nguyen2019robust,nguyen2019self,li2020dividemix}, relabeling \cite{yan2016robust,jiang2019hyperspectral,sahota2018energy}
    \end{tabular}\\\cline{2-2} 
    & \begin{tabular}[c]{@{}l@{}}\textbf{4. \nameref{samplechoosing}}\\ 
        a.\textit{\nameref{curriculumlearning}}: Screening loss \cite{han2018progressive}, teacher-student \cite{jiang2017mentornet}\\
        selecting uncertain samples \cite{chang2017active}, curriculum loss \cite{lyu2019curriculum}, data complexity \cite{guo2018curriculumnet}\\
        consistency with model \cite{reed2014training}\\    
        b.\textit{\nameref{multipleclassifiers}}: Consistency of networks \cite{malach2017decoupling}, co-teaching \cite{han2018co,yu2019disagreement,wang2019co,chen2019understanding}
    \end{tabular}\\\cline{2-2} 
    & \begin{tabular}[c]{@{}l@{}}\textbf{5. \nameref{sampleimportance}}\\ 
        Meta task \cite{ren2018learning,jenni2018deep,shu2019meta}, siamese network \cite{wang2018iterative}, pLOF \cite{xue2019robust}, abstention \cite{thulasidasan2019combating}\\
        estimate noise rate \cite{liu2015classification,wang2018multiclass}, similarity loss \cite{lee2018cleannet}, transfer learning \cite{litany2018soseleto}, $\theta$-distribution \cite{hu2019noise}
    \end{tabular}\\\cline{2-2} 
    & \begin{tabular}[c]{@{}l@{}}\textbf{6. \nameref{labelerquality}}\\ 
        EM \cite{raykar2009supervised,yan2014learning,khetan2017learning}, trace regularizer \cite{tanno2019learning}, crowd-layer \cite{rodrigues2018deep}, image difficulty estimate \cite{whitehill2009whose}\\
        consistency with network \cite{branson2017lean}, omitting probability variable \cite{izadinia2015deep}\\
        softmax layer per labeler \cite{guan2018said}
    \end{tabular}\\\hline

    \multirow{14}{*}{\rotatebox[origin=c]{90}{\textbf{\nameref{noisemodelfree}}}}  & 
    \begin{tabular}[c]{@{}l@{}}\textbf{1. \nameref{robustlosses}}\\ 
        Non-convex loss functions \cite{manwani2013noise,ghosh2015making,charoenphakdee2019symmetric}, 0-1 loss surrogate \cite{bartlett2006convexity}, MAE \cite{ghosh2017robust}, IMEA \cite{wang2019improved}\\
        Generalized cross-entropy \cite{zhang2018generalized}, symmetric loss \cite{wang2019symmetric}, unbiased estimator \cite{natarajan2013learning}, \\
        modified cross-entropy for omission \cite{mnih2012learning}, information theoric loss \cite{xu2019l_dmi}\\
        linear-odd losses \cite{patrini2016loss}, classification calibrated losses  \cite{van2015learning}, SGD with robust losses \cite{han2016convergence}
    \end{tabular}\\\cline{2-2} 
    & \begin{tabular}[c]{@{}l@{}}\textbf{2. \nameref{metalearning}}\\ 
        Choosing best methods \cite{garcia2016noise}, pumpout \cite{han2018pumpout}, noise tolerant parameter initialization \cite{junnan2018learning}, \\
        knowledge distillation \cite{li2017learning,kato2018improving}, gradient magnitude adjustment \cite{dehghani2017learning,dehghani2017avoiding}, meta soft labels \cite{algan2020meta}
    \end{tabular}\\\cline{2-2} 
    & \begin{tabular}[c]{@{}l@{}}\textbf{3. \nameref{regularizers}}\\ 
        Dropout \cite{srivastava2014dropout}, adversarial training \cite{goodfellow2014explaining}, mixup \cite{zhang2017mixup}, label smoothing \cite{pereyra2017regularizing,szegedy2016rethinking}\\
        pre-training \cite{hendrycks2019using}, dropout on final layer \cite{jindal2016learning}, checking dimensionality \\
        \cite{ma2018dimensionality}, auxiliary image regularizer \cite{azadi2015auxiliary}
    \end{tabular}\\\cline{2-2} 
    & \begin{tabular}[c]{@{}l@{}}\textbf{4. \nameref{ensemblemethods}}\\ 
        LogitBoost\&BrownBoost \cite{sun2011emprical}, noise detection based AdaBoost \cite{cao2012noise}, rBoost \cite{bootkrajang2013boosting}\\
        RBoost1\&RBoost2 \cite{miao2015rboost}, robust multi-class AdaBoost \cite{sun2016robust}
    \end{tabular}\\\cline{2-2} 
    & \begin{tabular}[c]{@{}l@{}}\textbf{5. \nameref{others}}\\ 
        Complementary labels \cite{yu2018learning,kim2019negative}, autoencoder reconstruction error \cite{xia2015learning}\\
        minimum covariance determinant \cite{lee2018robust}, less noisy data \cite{duan2017learning}, \\
        data quality \cite{choi2018choicenet}, prototype learning \cite{zhang2019metacleaner,seo2019combinatorial}, multiple instance learning \cite{niu2015visual,zhuang2017attend}
    \end{tabular}\\\hline 
    \end{tabular}
    \caption{Existing methods to deal with label noise in the literature}
    \label{table:methods}
\end{table*}
\end{singlespace}

	\section{Noise Model Based Methods} \label{noisemodelbased}
In the presence of noisy labels, the learning objective is to find the best estimator for hidden distribution $\mathcal{D}$, while iterating over distribution $\mathcal{D}_n$. If the mapping function $M:\mathcal{D} \rightarrow \mathcal{D}_n$ is known, it can be used to reverse the effect of noisy samples. Algorithms under this section simultaneously try to find underlying noise structure and train the base classifier with estimated noise parameters. They need a better estimate of $M$ to train better classifiers and better classifiers to estimate $M$ accurately. Therefore, they usually suffer from a chicken-egg problem. Approaches belonging to this category are explained in the following subsections.

\subsection{Noisy Channel} \label{noisychannel}
\begin{figure}[h]
    \centering
    \includegraphics[width=\columnwidth]{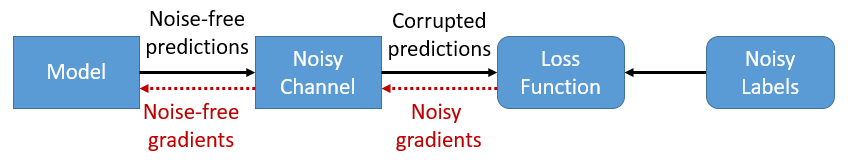}
    \caption{Noise can be modeled as a noisy channel on top of base classifier. Noisy channel adapts the characteristic of the noise so that base classifier is fed with noise-free gradients during traning.}
    \label{fig:noisychannel}
\end{figure}

The general setup for the noisy channel is illustrated in \autoref{fig:noisychannel}. Methods belonging to this category minimize the following risk

\begin{equation}
    \hat{R}_{l,\mathcal{D}}(f)=\dfrac{1}{N}\sum_{i=1}^{N}l(Q(f_\theta(x_i)),\tilde{y_i})
\end{equation}
where $Q(f_\theta(x_i)) = p(\tilde{y_i}|f_\theta(x_i))$ is the mapping from network predictions to given noisy labels. If $Q$ adapts the noise structure $p(\tilde{y}|y)$, then network will be forced to learn true mapping $p(y|x)$. 

$Q$ can be formulated with a \textit{noise transition matrix} $T$ so that $Q(f_\theta(x_i)) = Tf_\theta(x_i)$ where each element of the matrix represents the transition probability of given true label to noisy label, $T_{ij}=p(\tilde{y}=j|y=i)$. Since $T$ is composed of probabilities, weights coming from a single node should sum to one $\sum_{j}T_{ij}=1$. This procedure of correcting predictions to match given label distribution is also called \textit{loss-correction} \cite{patrini2017making}. 

A common problem in noisy channel estimation is scalability. As the number of classes increases, the size of the noise transition matrix increases exponentially, making it intractable to calculate. This can be partially avoided by allowing connections only among the most probable nodes \cite{goldberger2016training}, or predefined nodes \cite{han2018masking}. These restrictions are determined by human experts, which allows additional noise information to be inserted into the training procedure.  

The noisy channel is used only in the training phase. In the evaluation phase, the noisy channel is removed to get noise-free predictions of the base classifier. In these kinds of approaches, performance heavily depends on the accurate estimation of noisy channel parameters; therefore, works mainly focus on estimating $Q$. Various ways of formulating the noisy channel are explained below.

\subsubsection{Explicit calculation} \label{noisychannelexplicit} Noise transition matrix is calculated explicitly, and then the base classifier is trained using this matrix. Assuming dataset is balanced in terms of clean representative samples and noisy samples, so that there exists samples for each class with $p(y=\tilde{y}_i|x_i)=1$, \cite{patrini2017making} constructs $T$ just based on noisy class probability estimates of a pre-trained model, so-called \textit{confusion matrix}. A similar approach is followed in \cite{hendrycks2018using}; however, the noise transition matrix is calculated from the network's confusion matrix on the clean subset of data. Two datasets are gathered in \cite{chen2015webly}, namely: easy data and hard data. The classifier is first trained on the easy data to extract similarity relationships among classes. Afterward, the calculated similarity matrix is used as the noise transition matrix. Another method proposed in \cite{sukhbaatar2014training} calculates the confusion matrix on both noisy data and clean data. Then, the difference between these two confusion matrices gives $T$. 

\subsubsection{Iterative calculation} \label{noisychanneliterative} Noise transition matrix is estimated incrementally during the training of the base classifier. In \cite{bekker2016training,goldberger2016training} expectation-maximization (EM) \cite{dawid1979maximum} is used to iteratively train network to match given distribution and estimate noise transition matrix given the model prediction. The same approach is used on medical data with noisy labels in \cite{dgani2018training}. \cite{sukhbaatar2014training} adds a linear fully connected layer as a last layer of the base classifier, which is trained to adapt noise behavior. To avoid this additional layer to converge the identity matrix and base classifier overfitting the noise, the weight decay regularizer is applied to this layer. \cite{xia2019anchor} suggests using class probability estimates on anchor points (data points that belong to a specific class almost surely) to construct the noise transition matrix. In the absence of a noise-free subset of data, anchor points are extracted from data points with high noisy class posterior probabilities. Then, the matrix is updated iteratively to minimize loss during training. Instead of using softmax probabilities, \cite{yao2019safeguarded} models noise transition matrix in Bayesian form by projecting it into a Dirichlet-distributed space.

\subsubsection{Complex noisy channel}\label{noisychannelcomplex} Different then simple confusion matrix, some works formalize the noisy channel as a more complex function. This enables noisy channel parameters to be calculated not just by using network outputs but additional information about the content of data. For example, three types of label noises are defined in \cite{xiao2015learning}, namely: no noise, random noise, structured noise. An additional convolutional neural network (CNN) is used to interpret the noise type of each sample. Finally, the noisy layer aims to match predicted labels to noisy labels with the help of predicted noise type. Another work in \cite{misra2016seeing} proposes training an extra network as a relevance estimator, which attains the label's relevance to the given instance. Predicted labels are mapped to noisy labels with the consideration of relevance. If relevance is low, in case of noise, the classifier can still make predictions of true class and doesn't get penalized much for it. 

\subsection{Label Noise Cleaning} \label{labelnoisecleansing}
An obvious solution to noisy labels is to identify and correct suspicious labels to their corresponding true classes. Cleaning the whole dataset manually can be costly; therefore, some works propose to pick only suspicious samples to be sent to a human annotator to reduce the cost \cite{krause2016unreasonable}. However, this is still not a scalable approach. As a result, various algorithms are proposed in the literature. Including the label correction algorithm, the empirical risk takes the following form

\begin{equation}
    \hat{R}_{l,\mathcal{D}}(f)=\dfrac{1}{N}\sum_{i=1}^{N}l(f_\theta(x_i),G(\tilde{y_i},x_i))
\end{equation}
where $G(\tilde{y_i},x_i)=p(y_i|\tilde{y_i},x_i)$ represents the label cleaning algorithm. Label cleaning algorithms rely on a feature extractor to map data to the feature domain to investigate noisiness. While some works use a pre-trained network as the feature extractor, others use the base classifier as it gets more and more accurate during training. This approach results in an iterative framework: as the classifier gets better, the label cleaning is more accurate, and as the label quality gets better, the classifier gets better. From this point of view, label cleaning can be viewed as a dynamically evolving component of the system instead of preprocessing of the data. Such methods usually tackle the difficulty of distinguishing informative hard samples from those with noisy labels \cite{frenay2014classification}. As a result, they can end up removing too many samples or changing labels in a delusional way. Approaches for label cleaning can be separated according to their need for clean data or not.

\subsubsection{Using data with clean labels} \label{labelcleansingclean} In the existence of a clean subset of data, the aim is to fuse noise-free label structure to noisy labels for correction. If the clean subset is large enough to train a network, one obvious way is to relabel noisy labels by predictions of the network trained on clean data. For relabeling, \cite{lee2019photometric} uses alpha blending of given noisy labels and predicted labels. An ensemble of networks trained with different subsets of the dataset is used in \cite{yuan2018iterative}. If they all agree on the label, it is changed to the predicted label; otherwise, it is set to a random label. Instead of keeping the noisy label, setting it randomly helps break the noise structure and makes noise more uniformly distributed in label space. In \cite{vahdat2017toward} a graph-based approach is used, where a conditional random field extracts relation among noisy labels and clean labels.

\subsubsection{Using data with both clean and noisy labels} \label{labelcleansingcleannoisy} Some works rely on a subset of data, for which both clean and noisy labels are provided. Then label noise structure is extracted from these conflicting labels and used to correct noisy data. In \cite{veit2017learning}, the label cleaning network gets two inputs: extracted features of instances by the base classifier and corresponding noisy labels. Label cleaning network and base classifier are trained jointly so that label cleaning network learns to correct labels on the clean subset of data and provides corrected labels for base classifier on noisy data. Same approach is decoupled in \cite{dehghani2017fidelity} in teacher-student manner. First, the student is trained on noisy data. Then features are extracted from the clean data via the student model, and the teacher learns the structure of noise depending on these extracted features. Afterward, the teacher predicts soft labels for noisy data, and the student is again trained on these soft labels for fine-tuning.

\subsubsection{Using data with just noisy labels} \label{labelcleansingnoisy} Noise-free data is not always available, so the primary approach in this situation is to estimate cleaner posterior label distribution incrementally. However, there is a possible undesired solution to this approach so that all labels are attained to a single class and base network predicting constant class, which would result in delusional top training accuracy. Therefore, additional regularizers are commonly used to make label posterior distribution even. A joint optimization framework for both training base classifier and propagating noisy labels to cleaner labels is presented in \cite{tanaka2018joint}. Using expectation-maximization, both classifier parameters and label posterior distribution is estimated to minimize the loss. A similar approach is used in \cite{yi2019probabilistic} with additional compatibility loss conditioned on label posterior. Considering noisy labels are in the minority, this term assures posterior label distribution does not diverge too much from the given noisy label distribution so that majority of the clean label contribution is not lost. \cite{liu2017self,zheng2020error} deploy a confidence policy where labels are determined by either network output or given noisy labels, depending on the confidence of the model's prediction. Arguing that, in the case of noisy labels, the model first learns correctly labeled data and then overfits to noisy data, \cite{arazo2019unsupervised} aims to extract the probability of a sample being noisy or not from its loss value. To achieve this, the loss of each instance is fitted by a beta mixture model, which models the label noise in an unsupervised manner. \cite{zhang2017improving} proposes a two-level approach. In the first stage, with any chosen inference algorithm, the ground truth labels are determined, and data is divided into two subsets as noisy and clean. In the second stage, an ensemble of weak classifiers is trained on clean data to predict true labels of noisy data. Afterward, these two subsets of data are merged to create the final enhanced dataset. \cite{han2019deep} constructs prototypes that can represent deep feature distribution of the corresponding class for each class. Then corrected labels are found by checking similarity among data samples and prototypes. \cite{yao2018deep} introduces a new parameter, namely \textit{quality embedding}, which represents the trustworthiness of data. Depending on two latent variables, true class probability and quality embedding, an additional network tries to extract each instance's true class. In a multi-labeled dataset, where each instance has multiple labels representing its content, some labels may be partially missing resulting in partial labels. In the case of partial labels, \cite{durand2019learning} uses one network to find and estimate easy missing labels and another network to be trained on this corrected data. \cite{zhong2019graph} formulates video anomaly detection as a classification with label noise problem and trains a graph convolutional label noise cleaning network depending on features and temporal consistency of video snippets.

\subsection{Dataset Pruning} \label{datasetpruning}
Instead of correcting noisy labels to their true classes, an alternative approach is to remove them. While this would result in loss of information, preventing the harmful impact of noise may result in better performance. In these methods, there is a risk of removing too many samples. Therefore, it is crucial to remove as few samples as possible to prevent unnecessary data loss. 

There are two alternative ways for data pruning. The first option is to remove noisy samples completely and train the classifier on the pruned dataset. The second option is to remove just labels of noisy data and transform the dataset into two subsets as; labeled and unlabeled data. Then semi-supervised learning algorithms can be employed on the resultant dataset.

\subsubsection{Removing Data} The most straightforward approach is to remove instances misclassified by the base network \cite{delany2012profiling}. \cite{garcia2015using} uses an ensemble of filtering methods, where each of them assigns a noisiness level for each sample. Then, these predictions are combined, and data with the highest noisiness level predictions are removed. \cite{luengo2018cnc} extends this work with label correction. If the majority of noise filters predict the same label for the noisy instance, it's label is corrected to the predicted label. Otherwise, it is removed from the dataset. In \cite{northcutt2017learning}, with the help of a probabilistic classifier, training data is divided into two subsets: confidently clean and noisy. Noise rates are estimated according to the sizes of these subsets. Finally, relying on the base network's output confidence in data instances, the number of most unconfident samples is removed according to the estimated noise rate. In \cite{wu2018light}, transfer learning is used so that network trained on a clean dataset from a similar domain is fine-tuned on the noisy dataset for relabeling. Afterward, the network is again trained on relabeled data to re-sample the dataset to construct a final clean dataset. In \cite{huang2019o2u}, the learning rate is adjusted cyclicly to change network status between underfitting and overfitting. Since, while underfitted, noisy samples cause high loss, samples with large noise during this cyclic process are removed. \cite{sharma2020noiserank} first train network on noisy data and extract feature vectors by using this model. Afterward, data is clustered with the K-means algorithm running on extracted features, and outliers are removed. \cite{li2016noise} provides a comparison of performances of various noise-filtering methods for crowd-sourced datasets.

\subsubsection{Removing Labels} The simplest option is to employ straightforward semi-supervised training on labeled and unlabeled data \cite{ding2018semi}. Alternatively, label removing can be done iteratively in each epoch to update the dataset for better utilization of semi-supervised learning dynamically. \cite{nguyen2019robust} uses consistency among the given label and moving average of model predictions to evaluate if the provided label is noisy or not. Then the model is trained on clean samples on the next iteration. This procedure continues until convergence to the best estimator. The same approach is used in \cite{nguyen2019self} with a little tweak. Instead of comparing with given labels, the moving average of predictions is compared with predicted labels in the current epoch. To avoid the data selection biased caused by one model, \cite{li2020dividemix} uses two models to select an unlabeled set for each other. Afterward, each network is trained in a semi-supervised learning manner on the dataset chosen by its peer network. Another approach in this class is to train a network on labeled and unlabeled data and then use it to relabel noisy data \cite{sahota2018energy,yan2016robust}. Assuming that correctly labeled data account for the majority, \cite{jiang2019hyperspectral} proposes splitting datasets into labeled and unlabeled subgroups randomly. Then, labels are propagated to unlabeled data using a similarity index among instances. This procedure is repeated to produce multiple labels per instance, and then the final label is set with majority voting. 

\subsection{Sample Choosing} \label{samplechoosing}
A widely used approach to overcome label noise is to manipulate the input stream to the classifier. Guiding the network with choosing the right instances to feed can help the classifier finding its way easier in the presence of noisy labels. It can be formulated as follows

\begin{equation}
    \hat{R}_{l,\mathcal{D}}(f)=\dfrac{1}{N}\sum_{i=1}^{N}V(x_i,y_i)l(f_\theta(x_i),\tilde{y_i}))
    \label{eq:sc}
\end{equation}
where $V(x_i,y_i)\in\{0,1\}$ is a binary operator that decides to whether use the given data $(x_i,y_i)$ or not. If $V(x_i,y_i)=1$ for all data, then it turns out to be classical risk minimization (\autoref{eq:sc}). If $V$ happens to be a static function, which means choosing the same samples during whole training according to a predefined rule, then it turns out to be dataset pruning, as explained in \autoref{datasetpruning}. Differently, sample choosing methods continuously monitor the base classifier and select samples to be trained on for the next training iteration. The task can be seen as drawing a path through data that would mimic the noise-free distribution of $\mathcal{D}$. Since these methods operate outside of the existing system, they are easier to attach to the existing algorithm at hand by just manipulating the input stream. However, it is vital to keep the balance so that system does not ignore unnecessarily large quantities of data. Additionally, these methods prioritize low loss samples, which results in a slow learning rate since hard informative samples are considered only in the later stages of training. Two major approaches under this group are discussed in the following subsections.

\subsubsection{Curriculum Learning} \label{curriculumlearning} Curriculum learning (CL) \cite{bengio2009curriculum}, inspired from human cognition, proposes to start from easy samples and go through harder samples to guide training. This learning strategy is also called \textit{self-paced learning} \cite{kumar2010self,jiang2014self} when prior to sample hardness is not known and inferred from the loss of the current model on that sample. In the noisy label framework, clean labeled data can be accepted as an easy task, while noisily labeled data is the harder task. Therefore, the idea of CL can be transferred to label noise setup as starting from confidently clean instances and go through noisier samples as the classifier gets better. Various screening loss functions are proposed in \cite{han2018progressive} to sort instances according to their noisiness level. The teacher-student approach is implemented in \cite{jiang2017mentornet}, where the teacher's task is to choose confidently clean samples for the student. Instead of using a predefined curriculum, the teacher constantly updates its curriculum depending on the student's outputs. Arguing that CL slows down the learning speed, since it focuses on easy samples, \cite{chang2017active} suggests choosing uncertain samples that are mispredicted sometimes and correctly on others during training. These samples are assumed to be probably not noisy since noisy samples should be mispredicted all the time. Arguing that it is hard to optimize 0-1 loss, \textit{curriculum loss} that chooses samples with low loss values for loss calculation, is proposed as an upper bound for 0-1 loss in \cite{lyu2019curriculum}. In \cite{guo2018curriculumnet}, data is split into subgroups according to their complexities. Since less complex data groups are expected to have more clean labels, training will start from less complex data and go through more complex instances as the network gets better. Next samples to be trained on can be chosen by checking the consistency of the label with the network prediction. In \cite{reed2014training}, if both label and model prediction of the given sample is consistent, it is used in the training set. Otherwise, the model has a right to disagree. Iteratively this provides better training data and a better model. However, there is a risk of the model being too skeptical and choosing labels in a delusional way; therefore, consistency balance should be established. 

\subsubsection{Multiple Classifiers} \label{multipleclassifiers} Some works use multiple classifiers to help each other to choose the next batch of data to train on. This is different from the teacher-student approach since none of the networks supervise the other but rather help each other out. Multiple models can provide robustness since networks can correct each other's mistakes due to their differences in learned representations. For this setup to work, the initialization of the classifiers is essential. They are most likely to be initialized with a different subset of the data. If they have the same weight initializations, there is no update since they agree to disagree with labels. In \cite{malach2017decoupling} label is assumed to be noisy if both networks disagree with the given label, and update on model weights happens only when the prediction of two networks conflicts. The paradigm of \textit{co-teaching} is introduced in \cite{han2018co}, where two networks select the next batch of data for each other. The next batch is chosen as the data batch, which has small loss values according to the pair network. It is claimed that using one network accumulates the noise-related error, whereas two networks filter noise error more successfully. The idea of co-teaching is further improved by iterating over data where two networks disagree to prevent two networks converging each other with the increasing number of epochs \cite{yu2019disagreement,wang2019co}. Another work using co-teaching first trains two networks on a selected subset for a given number of epochs and then moves to the full dataset \cite{chen2019understanding}. 

\subsection{Sample Importance Weighting} \label{sampleimportance}
Similar to sample choosing, training can be made more effective by assigning weights to instances according to their estimated noisiness level. This has an effect of emphasizing cleaner instances for a better update on model weights. Following empirical risk is minimized by these algorithms.

\begin{equation}
    \hat{R}_{l,\mathcal{D}}(f)=\dfrac{1}{N}\sum_{i=1}^{N}\beta(x_i,y_i)l(f_\theta(x_i),\tilde{y_i}))
\end{equation}
where $\beta(x_i,y_i)$ determines the instance dependent weight. If $\beta$ would be binary, then formulation is the same with sample choosing, as explained in \autoref{samplechoosing}. Differently, here $\beta$ is not binary and has a different value for each instance. Like in sample choosing algorithms, $\beta$ is a dynamic function, which means weights for instances keep changing during the training. Therefore, it is commonly a challenge to prevent $\beta$ changing too rapidly and sharply, such that it disrupts the stabilized training loop. Moreover, these methods commonly suffer from accumulated errors. As a result, they can easily get biased towards a certain subset of data. There are various methods proposed to obtain optimal $\beta$ to fade away the negative effects of noise.

The simplest approach would be, in case of availability of both clean and noisy data, weighting clean data more \cite{sukhbaatar2014training}. However, this utilizes information poorly; moreover, clean data is not always available. Works of \cite{ren2018learning} and \cite{jenni2018deep}, uses meta-learning paradigm to determine the weighting factor. In each iteration, gradient descent step on the given mini-batch for weighting factor is performed so that it minimizes the loss on clean validation data. A similar method is adopted in \cite{shu2019meta}, but instead of implicit calculation of the weighting factor, multi layer perceptron (MLP) is used to estimate the weighting function. \textit{Open-set noisy labels}, where data samples associated with noisy labels might belong to a true class that is not present in the training data, are considered in \cite{wang2018iterative}. Siamese network is trained to detect noisy labels by learning discriminative features to apart clean and noisy data. Noisy samples are iteratively detected and pulled from clean samples. Then, each iteration weighting factor is recalculated for noisy samples, and the base classifier is trained on the whole dataset. \cite{xue2019robust} also iteratively separates noisy samples and clean samples. On top of that, not to miss valuable information from clean hard samples, noisy data are weighted according to their noisiness level, estimated by pLOF \cite{kriegel2009loop}. \cite{thulasidasan2019combating} introduces \textit{abstention}, which gives option to abstain samples, depending on their loss value, with an abstention penalty. Therefore, the network learns to abstain from confusing samples, and with the abstention penalty, the tendency to abstain can be adjusted. In \cite{liu2015classification}, weighting factor is conditioned on distribution of training data, $\beta(X,Y)= P_\mathcal{D}(X,Y)/P_{\mathcal{D}_n}(X,\tilde{Y})$. The same methodology is extended to the multi-class case in \cite{wang2018multiclass}. In \cite{lee2018cleannet}, the weighting factor is determined by checking instance similarity to its representative class prototype in the feature domain. \cite{litany2018soseleto} formulates the problem as transfer learning where the source domain is noisy data, and the target domain is a clean subset of data. Then weighting in the source domain is arranged in a way to minimize target domain loss. \cite{hu2019noise} uses $\theta$ values of samples in $\theta$-distribution to calculate their probability of being clean and use this information to weight clean samples more in training. 

\begin{figure}[h]
    \centering
    \includegraphics[width=\columnwidth]{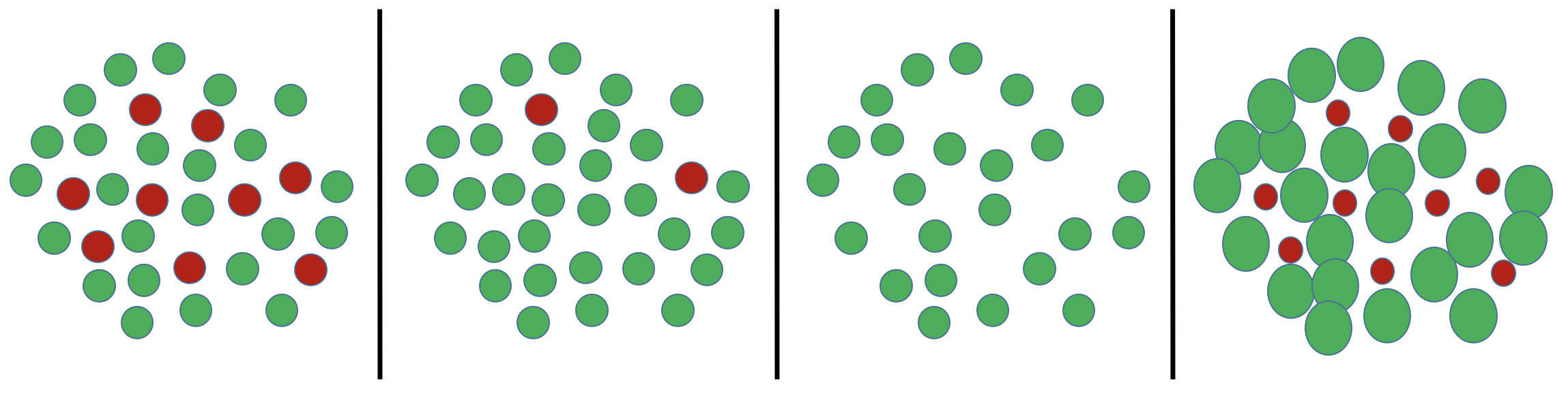}
    \caption{Illustration of different types of algorithms. Starting from left; 1) representation of samples from a single class in the 2d-space. Green samples represent the clean samples, and red ones represent noisy samples. 2) label noise cleaning algorithms aim to correct the labels of noisy data 3) dataset pruning methods aim to eliminate noisy data (or just their labels) 4) sample importance weighting algorithms aim to up-weight clean samples and down-weight noisy samples (which is illustrated by size)}
    \label{fig:methods}
\end{figure}

\subsection{Labeler Quality Assessment} \label{labelerquality}
As explained in \autoref{sourcesoflabelnoise}, there can be several reasons for the dataset to be labeled by multiple annotators. Each labeler may have a different level of expertise, and their labels may occasionally contradict each other. This is a typical case in crowd-sourced data \cite{vuurens2011much,wais2010towards,ipeirotis2010quality} or datasets that require a high level of expertise such as medical imaging \cite{pechenizkiy2006class}. Therefore, modeling and using labeler characteristics can significantly increase performance \cite{guan2018said}.

In this setup, there are two unknowns; noisy labeler characteristics and ground truth labels. One can estimate both with expectation-maximization \cite{raykar2009supervised,yan2014learning,khetan2017learning}. If noise is assumed to be y-dependent, the labeler characteristic can be modeled with a noise transition matrix, just like in \autoref{noisychannel}. \cite{tanno2019learning} adds a regularizer to the loss function, which is the sum of traces of annotator confusion matrices, to force sparsity on each labeler's estimated confusion matrix. A similar approach is implemented in \cite{rodrigues2018deep}, where a crowd-layer is added to the end of the network. In \cite{whitehill2009whose}, xy-dependent noise is also considered by taking image complexities into account. Human annotators and computer vision systems are used mutually in \cite{branson2017lean}, where consistency among predictions of these two components is used to evaluate labelers' reliability. \cite{izadinia2015deep} deals with the noise when the labeler omits a tag in the image. Therefore, instead of the noise transition matrix for labelers, the omitting probability variable is used, which is estimated together with the true class using the expectation-maximization algorithm. Separate softmax layers are trained for each annotator in \cite{guan2018said} and an additional network to predict the true class of data depending on the outputs of labeler specific networks and features of data. This setup enables to model each labeler and their overall noise structure in separate networks. 

\subsection{Discussion}
A visual illustration of some of the methods is presented in \autoref{fig:methods}. Noise model based methods are heavily dependent on the accurate estimate of the noise structure. This brings a dilemma. For a better noise model, one needs better estimators, and for better estimators, it is necessary to have a better estimate of underlying noise. Therefore, many approaches can be seen as an expectation-maximization of both noise estimation and classification. However, it is essential to prevent the system diverging from reality. Therefore regularizing noise estimates and not letting it getting delusional is crucial. To achieve this, the literature commonly makes assumptions about the underlying noise structure, which damages their applicability to different setups. On the other hand, this lets any prior information about the noise be inserted into the system for a head-start. It is also useful to handle domain-specific noise. One another advantage of these algorithms is they decouple noise estimation and classification tasks. Therefore, they are easier to implement on an existing classification algorithm at hand. 
	\section{Noise Model Free Methods} \label{noisemodelfree}
These methods aim to achieve label noise robustness without explicitly modeling it, but rather designing robustness in the proposed algorithm. Noisy data is treated as an anomaly; therefore, these methods are similar to overfit avoidance. They commonly rely on the classifier's internal noise tolerance and aim to boost performance by regularizing undesired memorization of noisy data. Various methodologies are presented in the following subsections.

\subsection{Robust Losses} \label{robustlosses}
A loss function is said to be noise robust if the classifier learned with noisy and noise-free data achieves the same classification accuracy \cite{manwani2013noise}. Algorithms under this section aim to design loss function so that the existence of the noise would not decrease the performance. However, it is shown that noise can badly affect the performance even for the robust loss functions \cite{frenay2014classification}. Moreover, these methods treat both noisy and clean data in the same way, which prevents the utilization of any prior information over data distribution.

In \cite{manwani2013noise}, it is shown that certain non-convex loss functions, such as 0-1 loss, has noise tolerance much more than commonly used convex losses. Extending this work \cite{ghosh2015making,charoenphakdee2019symmetric} derives sufficient conditions for a loss function to be noise tolerant for uniform noise. Their work shows that, if the given loss function satisfies $\sum_{k}l(f_\theta(x),k) = C, \forall x \in X$ where $C$ is a constant value, then loss function is tolerant to uniform noise. In this content, they empirically show that none of the standard convex loss functions has noise robustness while 0-1 loss has, up to a certain noise ratio. However, 0-1 loss is non-convex and non-differentiable; therefore, surrogate loss of 0-1 loss is proposed in \cite{bartlett2006convexity}, which is still noise sensitive. Widely used \textit{categorical cross entropy (CCE)} loss is compared with \textit{mean absolute value of error (MAE)} in the work of \cite{ghosh2017robust}, where it is shown empirically that mean absolute value of error is more noise tolerant. \cite{wang2019improved} shows that the robustness of MEA is due to its weighting scheme. While CCE is sensitive to abnormal samples and produces bigger gradients in magnitude, MAE treats all data points equally, which would result in an underfitting of data. Therefore, \textit{Improved mean absolute value of error (IMAE)}, which is an improved version of MAE, is proposed in \cite{wang2019improved}, where gradients are scaled with a hyper-parameter to adjusts weighting variance of MAE. \cite{zhang2018generalized} also argues that MAE provides a much lower learning rate than CCE; therefore, a new loss function is suggested, which combines the robustness of MAE and implicit weighting of CCE. With a tuning parameter, the loss function characteristic can be adjusted in a line from MAE to CCE. Loss functions are commonly not symmetric, meaning that $l(f_\theta(x_i),y_i) \neq l(y_i,f_\theta(x_i))$. Inspired from the idea of symmetric KL-divergence, \cite{wang2019symmetric} proposes symmetric cross entropy loss $l_{SCE}(f_\theta(x_i),y_i) = l(f_\theta(x_i),y_i) + l(y_i,f_\theta(x_i))$ to battle noisy labels. 

Given that noise prior is known, \cite{natarajan2013learning} provides two surrogate loss functions using the prior information about label noise, namely, an unbiased and a weighted estimator of the loss function. \cite{mnih2012learning} considers asymmetric omission noise for the binary classification case, where the task is to find road pixels from a satellite map image. Omission noise makes the network less confident about its predictions, so they modified cross-entropy loss to penalize the network less for producing wrong but confident predictions since these labels are more likely to be noisy. Instead of using distance-based loss, \cite{xu2019l_dmi} proposes to use information-theoretic loss, in which determinant based mutual information \cite{kong2020dominantly} between given labels and predictions are evaluated for loss calculation. Weakly supervised learning with noisy labels are considered in \cite{patrini2016loss}, and necessary conditions for loss to be noise tolerant are drawn. \cite{van2015learning} shows that classification-calibrated loss functions are asymptotically robust to symmetric label noise. Stochastic gradient descent with robust losses is analyzed in general \cite{han2016convergence} and shown to be more robust to label noise than its counterparts.

\subsection{Meta Learning} \label{metalearning}
With the recent advancements of deep neural networks, the necessity of hand-designed features for computer vision systems are mostly eliminated. Instead, these features are learned autonomously via machine learning techniques. Even though these algorithms can learn complex functions on their own, there remain many hand-designed parameters such as network architecture, loss function, optimizer algorithm, and so on. Meta-learning aims to eliminate these necessities by learning not just the required complex function for the task but also learning the learning itself \cite{andrychowicz2016learning,finn2017model}. Algorithms belonging to this category usually implement an additional learning loop for the meta objective, optimizing the base learning procedure. In general, the biggest drawback of these methods is their computational cost. Since they require nested loops of gradient computations for each training loop, they are several times slower than the conventional training process.

Designing a task beyond classical supervised learning in a meta-learning fashion has been used to deal with label noise as well. A meta task is defined as predicting the most suitable method, among the family of methods, for a given noisy dataset in \cite{garcia2016noise}. \textit{Pumpout} \cite{han2018pumpout} presents a meta objective as recovering the damage done by noisy samples by erasing their effect on model via \textit{scaled gradient ascent}. As a meta-learning paradigm, model-agnostic-meta-learning (MAML) \cite{finn2017model} seeks optimal weight initialization that can easily be fine-tuned for the desired objective. A similar mentality is used in \cite{junnan2018learning} for noisy labels, which aims to find noise-tolerant model parameters that are less prone to noise under teacher-student training framework \cite{rasmus2015semi, tarvainen2017mean}. Multiple student networks are fed with data corrupted by synthetic noise. A meta objective is defined to maximize consistency with teacher outputs obtained from raw data without synthetic noise. Therefore, student networks are forced to find most noise robust weight initialization such that weight update will still be consistent after training an epoch on synthetically corrupted data. Then, final classifier weights are set as an exponential moving average of student networks. 

Alternatively, in the case of available clean data, a meta objective can be defined to utilize this information. The approach used in \cite{li2017learning} is to train a teacher network in a clean dataset and transfer its knowledge to the student network to guide the training process in the presence of mislabeled data. They used \textit{distillation} technique proposed in \cite{hinton2015distilling} for controlled transfer of knowledge from teacher to student. A similar methodology of using \textit{distillation} and label correction in the human pose estimation task is implemented in \cite{kato2018improving}. In \cite{dehghani2017learning,dehghani2017avoiding}, the target network is trained on excessive noisy data, and the confidence network is trained on a clean subset. Inspiring from \cite{andrychowicz2016learning}, the confidence network's task is to control the magnitude of gradient updates to the target network so that noisy labels are not resulting in updating gradients. \cite{algan2020meta} uses clean data to produce soft labels for noisy data, for which the classifier trained on it would give the best performance on the clean data. As a result, it seeks optimal label distribution to provide the most noise robust learning for the base classifier.

\subsection{Regularizers} \label{regularizers}
Regularizers are well known to prevent DNNs from overfitting noisy labels. From this perspective, these methods treat performance degradation due to noisy data as overfitting to noise. Even though this assumption is mostly valid in random noise, it may not be the case for more complex noises. Some widely used techniques are weight decay, dropout \cite{srivastava2014dropout}, adversarial training \cite{goodfellow2014explaining}, mixup \cite{zhang2017mixup}, label smoothing \cite{pereyra2017regularizing,szegedy2016rethinking}. \cite{hendrycks2019using} shows that pre-training has a regularization effect in the presence of noisy labels. In \cite{jindal2016learning} an additional softmax layer is added, and dropout regularization is applied to this layer, arguing that it provides more robust training and prevents memorizing noise due to randomness of dropout \cite{srivastava2014dropout}. \cite{ma2018dimensionality} proposes a complexity measure to understand if the network starts to overfit. It is shown that learning consists of two steps: 1) dimensionality compression, which models low-dimensional subspaces that closely match the underlying data distribution, 2) dimensionality expansion, which steadily increases subspace dimensionality to overfit the data. The key is to stop before the second step. \textit{Local intrinsic dimensionality} \cite{houle2013dimensionality} is used to measure the complexity of the trained model and stop before it starts to overfit. \cite{azadi2015auxiliary} takes a pre-trained network on a different domain and fine-tunes it for the noisy labeled dataset. Groups of image features are formed, and group sparsity regularization is imposed so that model is forced to choose relative features and up-weights the reliable images.

\subsection{Ensemble Methods} \label{ensemblemethods}
It is well known that bagging is more robust to label noise than boosting \cite{dietterich2000experimental}. Boosting algorithms like AdaBoost puts too much weight on noisy samples, resulting in overfitting the noise. However, the degree of label noise robustness changes for the chosen boosting algorithm. For example, it is shown that BrownBoost and LogitBoost are more robust than AdaBoost \cite{sun2011emprical}. Therefore, noise-robust alternatives of AdaBoost is proposed in literature, such as noise detection based AdaBoost \cite{cao2012noise}, rBoost \cite{bootkrajang2013boosting}, RBoost1\&RBoost2 \cite{miao2015rboost} and robust multi-class AdaBoost \cite{sun2016robust}.

\subsection{Others} \label{others}
\textit{Complementary labels} define classes that observations do not belong to. For example, in the case of ten classes, there is one true class and nine complimentary classes for each instance. Since annotators are less likely to mislabel, some works propose to work in complementary label space \cite{yu2018learning,kim2019negative}. \cite{xia2015learning} uses reconstruction error of autoencoder to discriminate noisy data from clean data, arguing that noisy data tend to have bigger reconstruction error. In \cite{lee2018robust}, the base model is trained with noisy data. An additional generative classifier is trained on top of the feature space generated by the base model. By estimating its parameters with \textit{minimum covariance determinant}, noise-robust decision boundaries are aimed to be found. In \cite{duan2017learning}, a special setup is considered where dataset consists of noisy and \textit{less-noisy} data for binary classification task. \cite{choi2018choicenet} aims to extract the quality of data instances. Assuming that the training dataset is generated from a mixture of the target distribution and other unknown distributions, it estimates the quality of data samples by checking the consistency between generated and target distributions. 

\textit{Prototype learning} aims to construct prototypes that can represent features of a class in order to learn clean representations. Some works in the literature \cite{zhang2019metacleaner,seo2019combinatorial} propose to create clean representative prototypes for noisy data, so that base classifier can be trained on them instead of noisy labels.

In multiple-instance learning, data are grouped in clusters, called bags, and each bag is labeled as positive if there is at least one positive instance in it and negative otherwise. The network is fed with a group of data and produces a single prediction for each bag by learning the inner discriminative representation of data. Since the group of images is used and one prediction is made, the existence of noisy labels along with true labels in a bag has less impact on learning. In \cite{niu2015visual}, authors propose to effectively choose training samples from each bag by minimizing the total bag level loss. Extra model is trained in \cite{zhuang2017attend} as an attention model, which determines parts of the images to be focused on. The aim is to focus on a few regions on correctly labeled images and not focus on any region for mislabeled images.

\subsection{Discussion}
Overall, methods belonging to this category treat noisy data as an anomaly. Therefore, they are in a similar line with overfit avoidance and anomaly detection. Even though this assumption may be quite valid for random noise, it loses its validity in the case of more complicated and structured noises. Since noise modeling is not decoupled from classification task explicitly, proposed methods are, in the general sense, embedded into the existing algorithm. This prevents their quick deployment to the existing system at hand. Moreover, algorithms belonging to meta-learning and ensemble methods can be computationally costly since they require multiple iterations of training loops.

	\section{Experiments} \label{experiments}
This section discusses how the proposed algorithms from the literature conduct experiments to test their robustness against label noise. In general, for quick implementation and testing, most of the works start by testing on toy datasets (MNIST\cite{lecun1998mnist}, MNIST-Fashion\cite{xiao2017fashion}, CIFAR10\&CIFAR100\cite{torralba200880}) with synthetic label noise. However, as explained in \autoref{labelnoisemodels}, there are various types of artificial noises. Moreover, each work experiment with a different model architecture and hyper-parameter set. Therefore, results on these datasets are not appropriate for a fair comparison of the algorithms. They are instead used as a proof of concept for the proposed algorithm.

Some works from the literature use two alternative datasets. The first one is the Food101N dataset \cite{lee2018cleannet} containing 310k images of food recipes belonging to 101 different classes. However, its noise ratio is pretty small (around 20\%), making it inadequate to evaluate noise robust algorithms' performance. The second option is the WebVision dataset \cite{li2017webvision} containing 2.4 million images crawled from Flickr website and Google Images search. This is a big dataset, which requires a lot of computational power to run algorithms. Some works conduct tests on this dataset by using data only from the first 50 classes, aiming to make it computationally feasible. But still, WebVision fails to provide a benchmarking dataset for the evaluation of noise robust algorithms.

To fill the absence of a benchmarking dataset, \cite{xiao2015learning} collects a large amount of images from the web with labels interpreted from the surrounding user tags. As a result, it has real-world noisy labels with an estimated noise ratio of around 40\%. Dataset consists of one million images belonging to 14 different classes. 50K, 14K and 10K additional images with verified clean labels for train, validation and test purposes. We observed a high majority of the methods do not use additional 50K clean data for training. Furthermore, the literature seems to have a consensus on the experimental setup. All methods use the same model architecture of ResNet50 \cite{he2016deep} with pre-trained parameters on Imagenet \cite{deng2009imagenet} and stochastic gradient descent optimizer. Considering the identical experimental setup and real-world noisiness of the dataset, the Clothing1M dataset is widely accepted as a benchmarking dataset.

We listed (to the best of our knowledge) all of the results presented on this dataset in \autoref{table:clothing1m}. We collected results only from the works trained on 1M noisy training data without additional 50K clean data for a fair evaluation. We sorted algorithms according to their test accuracy. Nevertheless, it should be noted that each method has its pros and cons, such as computational cost, memory requirements, etc.

\begin{table}[h]
    \centering
    \begin{tabular}{|l|l|l|}
        \hline \multicolumn{1}{|c|}{\textbf{Method}}    & \multicolumn{1}{c|}{\textbf{Category}}    & \multicolumn{1}{c|}{\textbf{Accuracy}}   \\ \hline
        \cite{algan2020meta}                            & \nameref{metalearning}                    & 76.02 \\ 
        \cite{li2020dividemix}                          & \nameref{datasetpruning}                  & 74.76 \\ 
        \cite{lee2018cleannet}                          & \nameref{sampleimportance}                & 74.69 \\ 
        \cite{han2019deep}                              & \nameref{labelnoisecleansing}             & 74.45 \\ 
        \cite{xia2019anchor}                            & \nameref{noisychannel}                    & 74.18 \\ 
        \cite{sharma2020noiserank}                      & \nameref{datasetpruning}                  & 73.77 \\ 
        \cite{shu2019meta}                              & \nameref{sampleimportance}                & 73.72 \\ 
        \cite{yi2019probabilistic}                      & \nameref{labelnoisecleansing}             & 73.49 \\ 
        \cite{junnan2018learning}                       & \nameref{metalearning}                    & 73.47 \\ 
        \cite{wang2019improved}                         & \nameref{robustlosses}                    & 73.20 \\ 
        \cite{yao2019safeguarded}                       & \nameref{noisychannel}                    & 73.07 \\ 
        \cite{zhang2019metacleaner}                     & \nameref{others}                          & 72.50 \\ 
        \cite{xu2019l_dmi}                              & \nameref{robustlosses}                    & 72.46 \\ 
        \cite{tanaka2018joint}                          & \nameref{labelnoisecleansing}             & 72.23 \\ 
        \cite{zheng2020error}                           & \nameref{labelnoisecleansing}             & 71.74 \\ 
        \cite{han2018masking}                           & \nameref{noisychannel}                    & 71.10 \\ 
        \cite{wang2019symmetric}                        & \nameref{robustlosses}                    & 71.02 \\ 
        \cite{arazo2019unsupervised}                    & \nameref{labelnoisecleansing}             & 71.00 \\ \hline
    \end{tabular}
    \caption{Leaderboard for algorithms tested on the Clothing1M dataset. All results are taken from the corresponding paper. For fair evaluation only the works which did not used additional 50k clean training data are presented.}
    \label{table:clothing1m}
\end{table}
	\section{Conclusion} \label{conclusion}
Throughout this paper, it is shown that label noise is an important obstacle to deal with in order to achieve desirable performance from real-world datasets. Despite its importance for supervised learning in practical applications, it is also an important step to collect datasets from the web \cite{yu2015lsun, zhou2017places}, design networks that can learn from unlimited web data with no human supervision \cite{fergus2010learning,chen2013neil,divvala2014learning,joulin2016learning}. Furthermore, beside image classification, there are more fields where dealing with mislabeled instances is important, such as generative networks \cite{kaneko2018label,thekumparampil2018robustness}, semantic segmentation \cite{lu2016learning,zhu2018improving,acuna2019devil}, sound classification \cite{fonseca2019learning} and more. All these factors make dealing with label noise an important step through self-sustained learning systems.

Different approaches to come through noisy label phenomenon are proposed in the literature. All methods have their advantages and disadvantages, so one can choose the most appropriate algorithm for the use case. However, in order to draw a generic line, we make the following suggestions. If the noise structure is domain-specific and there is prior information or assumption about its structure, noise model based methods are more appropriate. Among these models, one can choose the best-suited method according to need. For example, if noise can be represented as a noise transition matrix, noisy channel or labeler quality assessment for multi labeler case can be chosen. If the purpose is to purify the dataset as a preprocessing stage, then dataset pruning or label noise cleansing methods can be employed. Sample choosing or sample importance weighting algorithms are handy if instances can be ranked according to their informativeness on training. Unlike noise model-based algorithms, noise model free methods do not depend on any prior information about the noise structure. Therefore, they are easier to implement if noise is assumed to be random, and performance degradation is due to overfitting since they do not require the hassle of implementing an external algorithm for noise structure estimation. If there is no clean subset of data, robust losses or regularizers are appropriate options since they treat all samples the same. Meta-learning techniques can be used in the presence of a clean subset of data since they can easily be adapted to utilize this subset. 

Even though an extensive amount of research is conducted for machine learning techniques \cite{frenay2014classification}, deep learning in the presence of noisy labels is certainly an understudied problem. Considering its dramatic effect on DNNs \cite{zhang2016understanding}, there still are many open research topics in the field. For example, truly understanding the impact of label noise on deep networks can be a fruitful future research topic. \cite{zeiler2014visualizing} shows that each layer of CNN learns to extract different features from the data. Moreover, learned representations form a hierarchical pattern, where each layer learns more complex features from the previous layer. A fully connected layer uses features from the last layer to interpret the corresponding label on the final layer. Understanding which parts of the network is highly affected by label noise may help analyze the adverse effect of the label noise on neural networks. For example, if initial layers are affected, one can conclude that learned primitive features are corrupted, so the rest of the network cannot be adequately trained. On the other hand, if final convolution layers are affected, it can be said that the network can not form the hierarchical feature pattern throughout the convolutional layers. Alternatively, if the convolutional layers are not affected but the fully connected layer is the cause of the problem, it can be concluded that feature representation learning is not corrupted, but the network cannot correctly interpret meanings from the extracted features. Moreover, it can be interesting to investigate the cause of the problem for different types of label noise models presented in \autoref{labelnoisemodels}. If for different noise models, different parts of the network are affected, one can analyze the true nature of the label noise in the dataset by checking corruption in the neural network layers.

Alternatively, the question of how to train in the existence of both attribute and label noise is an understudied problem with significant potential for practical applications \cite{lee2019photometric}. \cite{wang2019symmetric} shows noisy labels degrades the learning, especially for challenging samples. So, instead of overfitting to noisy samples, underfitting to challenging samples may be the reason for the performance degradation, which is an open question to be answered in the future. Another possible research direction may be on the effort of breaking the structure of the noise to make it uniformly distributed in the feature domain \cite{jiang2019hyperspectral}. This approach would be handy where labelers have a particular bias.

A widely used approach for quick testing of proposed algorithms is to create noisy datasets by adding synthetic label noise to benchmarking toy datasets \cite{lecun1998mnist,xiao2017fashion,torralba200880,krizhevsky2009learning,netzer2011reading}. However, this prevents fair comparison and evaluation of algorithms since each work adds its own noise type. Some large datasets with noisy labels are proposed in literature \cite{rolnick2017deep,lee2018cleannet,li2017webvision,thomee2015new}. These datasets are collected from the web, and labels are attained from noisy user tags. Even though these datasets provide a useful domain for benchmarking proposed solutions, their noise rates are mostly unknown and they are biased in terms of data distribution for classes. Moreover, one can not adjust the noise rate for testing under extreme or moderate conditions. From this perspective, we believe literature lacks a noisy dataset where a major part of it has both noisy and verified labels; thus, the noise rate can be adjusted as desired.

Minimal attention is given to the learning from a noisy labeled dataset when there is a small amount of data. This can be a fruitful research direction considering its potential in fields where harvesting dataset is costly. For example, in medical imaging, collecting a cleanly annotated large dataset is not feasible most of the time \cite{lee2019photometric}, due to its cost or privacy. Effectively learning from a small amount of noisy data with no ground truth can significantly improve autonomous medical diagnosis systems. Even though some pioneer researches are available \cite{guan2018said,dgani2018training,xue2019robust}, there is still much more to be explored.

The ability to effectively learn from noisily labeled data brings up big opportunities for practical applications of machine learning algorithms. The bottleneck of data collection can easily be resolved with the massive amount of data collected from the web. Labels for this data can be assigned with simple algorithms (such as interpreting from the surrounding text \cite{xiao2015learning}). By effectively dealing with noisy labels, deep learning algorithms can be fed with massive datasets. Moreover, there are research opportunities for alternative usage of semi-supervised learning algorithms along with noisily labeled data. Common usage of semi-supervised learning methods for noisily labeled data is to remove the labels of noisy data and then train with conventional semi-supervised learning methods. Alternatively, algorithms can be developed to use three types of data; cleanly labeled data, noisily labeled data, and unlabeled data. With the help of these algorithms, a massive amount of unlabeled and noisy data can be effectively used under the supervision of a small cleanly annotated data. 

Besides classification, knowledge of learning from noisily labeled data can be used in alternative fields by transforming the task. For example, in a multi-labeled dataset, where each instance belongs to multiple classes, not all classes are equally relevant. One can assume labels with a small resemblance to the data sample as noisy and employ algorithms designed for learning from noisy labels \cite{lin2018attribute}. Similarly, \cite{fan2017complex} divides an untrimmed video into smaller parts and aims to find the video snippet most relevant to the video tag. Irrelevant video parts are assumed to be noisily labeled. Even though the original dataset does not have noisy labels, it can be transformed to use algorithms from the field. As a result, learning from noisy labels has many potentials in various areas besides straight image classification.

	\bibliography{surveyrefs} 

\end{document}